# Text Line Segmentation of Historical Documents: a Survey

Laurence Likforman-Sulem*, Abderrazak Zahour**, Bruno Taconet**

**GET-Ecole Nationale Supérieure des Télécommunications/TSI and CNRS-LTCI, 46 rue Barrault, 75013 Paris, France*

email: likforman@tsi.enst.fr

Phone: +33 1 45 81 73 28

Fax: +33 1 45 81 37 94

http://www.tsi.enst.fr/~lauli/

*** IUT, Université du Havre/GED, Place Robert Schuman, 76610 Le Havre, France*

email :{taconet|zahour}@univ-lehavre.fr

## Abstract

There is a huge amount of historical documents in libraries and in various National Archives that have not been exploited electronically. Although automatic reading of complete pages remains, in most cases, a long-term objective, tasks such as word spotting, text/image alignment, authentication and extraction of specific fields are in use today. For all these tasks, a major step is document segmentation into text lines. Because of the low quality and the complexity of these documents (background noise, artifacts due to aging, interfering lines), automatic text line segmentation remains an open research field. The objective of this paper is to present a survey of existing methods, developed during the last decade, and dedicated to documents of historical interest.

# 1. Introduction

Text line extraction is generally seen as a preprocessing step for tasks such as document structure extraction, printed character or handwriting recognition. Many techniques have been developed for page segmentation of printed documents (newspapers, scientific journals, magazines, business letters) produced with modern editing tools [57] [38] [14] [39] [2]. The segmentation of handwritten documents has also been addressed with the segmentation of address blocks on envelopes and mail pieces [9] [10] [15][48], and for authentication or recognition purposes [53] [60]. More recently, the development of handwritten text databases (IAM database, [34]) provides new material for handwritten page segmentation.

Ancient and historical documents, printed or handwritten, strongly differ from the documents mentioned above because layout formatting requirements were looser. Their physical structure is thus harder to extract. In addition, historical documents are of low quality, due to aging or faint typing. They include various disturbing elements such as holes, spots, writing from the verso appearing on the recto, ornamentation, or seals. Handwritten pages include narrow spaced lines with overlapping and touching components. Characters and words have unusual and varying shapes, depending on the writer, the period and the place concerned. The vocabulary is also large and may include unusual names and words. Full text recognition is in most cases not yet available, except for printed documents for which dedicated OCR can be developed.

However, invaluable collections of historical documents are already digitized and indexed for consulting, exchange and distant access purposes which protect them from direct manipulation. In some cases, highly structured editions have been established by scholars. But a huge amount of documents are still to be exploited electronically. To produce an electronic searchable form, a document has to be indexed. The simplest way of indexing a document consists in attaching its main characteristics such as date, place and author (the so called ‘metadata’). Indexing can be enhanced when the document structure and content are exploited. When a transcription (published version, diplomatic transcription) is available, it can be attached to the digitized document: this allows users to retrieve documents from textual queries.

Since text based representations do not reflect the graphical features of such documents, a better representation is obtained by linking the transcription to the document image. A direct correspondence can then be established between the document image and its content by text/image alignment techniques [55]. This allows the creation of indexes where the position of each word can be recorded, and of links between both representations. Clicking on a word on the transcription or in the index through a GUI allows users to visualize the corresponding image area and vice versa. To make such queries possible for handwritten sources of literary works, several projects have been carried out under EU and National Programs: for instance the so-called ‘philological workstation’ *Bambi* [6][8] and within the *Philectre* reading and editing environment [47]. The document analysis embedded in such systems provides tools to search for blocks, lines and words, and may include a dedicated handwriting recognition system. Interactive tools are generally offered for segmentation and recognition correction purposes. Several projects also concern printed material: *Debora* [5] and *Memorial* [3]. Partial or complete logical structure can also be extracted by document analysis and corrected with GUI as in the *Viadocs* project [11][18]. However, document structure can also be used when no transcription is available. Word spotting techniques [22] [55] [46] can retrieve similar words in the image document through an image query. When words of the image document are extracted by top down segmentation, which is generally the case, text lines are extracted first.

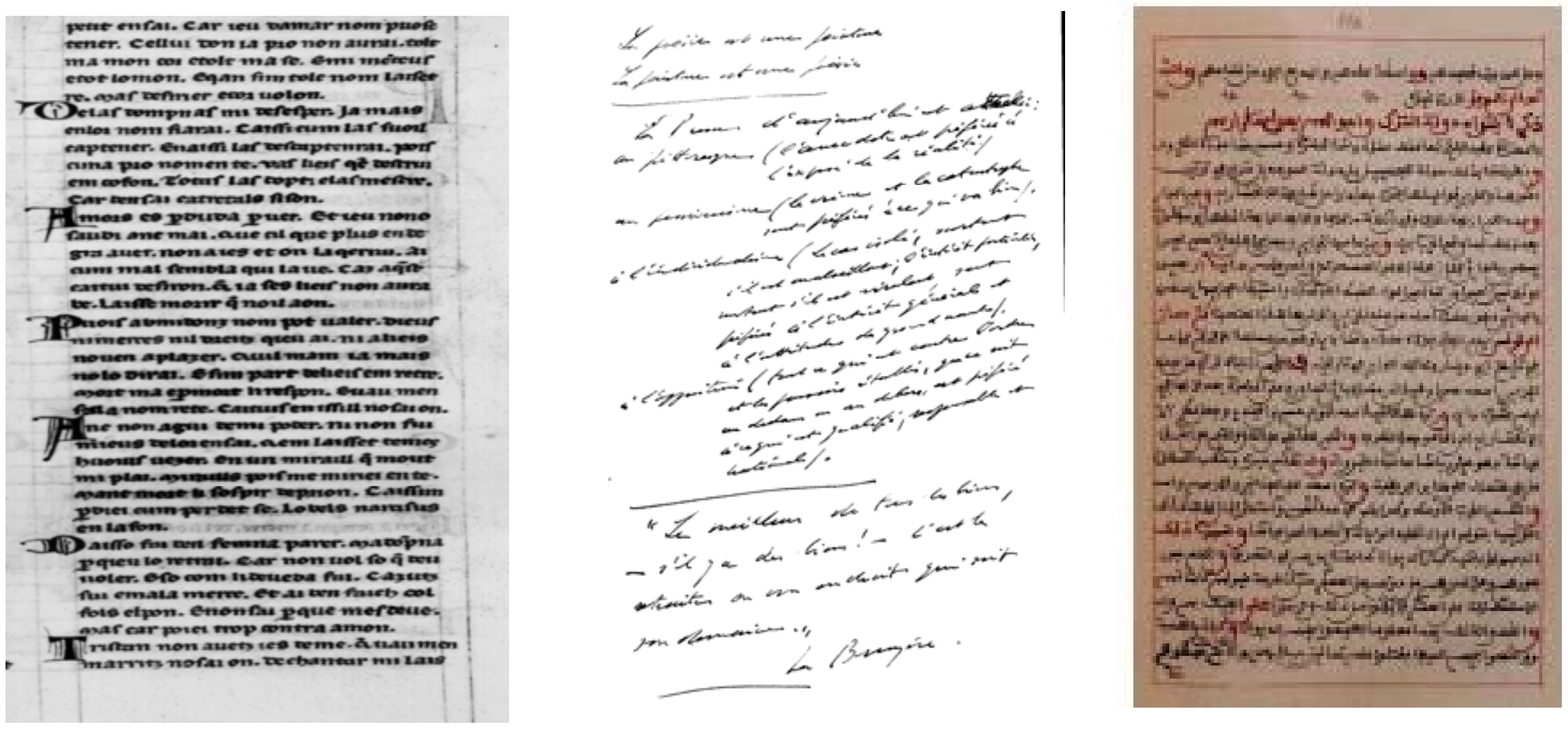

Fig. 1. Examples of historical documents a) Provencal medieval manuscript. b) one page from De Gaulle's diaries c) an ancient Arabic document from Tunisian Archives.

The authentication of manuscripts in the paleographic sense can also make use of document analysis and text line extraction. Authentication consists in retrieving writer characteristics independently from document content. It generally consists in dating documents, localizing the place where the document was produced, identifying the writer by using characteristics and features extracted from blank spaces, line orientations and fluctuations, word or character shapes [43] [27] [4].

Page segmentation into text lines is performed in most tasks mentioned above and overall performance strongly relies on the quality of this process. The purpose of this article is to survey the efforts made for historical documents on the text line segmentation task. Section 2 describes the characteristics of text line structures in historical documents and the different ways of defining a text line. Preprocessing of document images (gray level, color or black and white) is often necessary before text line extracting to prune superfluous information (non textual elements, textual elements from the verso) or to correctly binarize the image. This problem is addressed in Section 3.1. In Sections 3.2-3.7 we survey the different approaches to segment the clean image into text lines. A taxonomy is proposed, listed as projection profiles, smearing, grouping, Hough-based, repulsive-attractive network and stochastic methods. The majority of these techniques have been developed for the projects on historical documents mentioned above. In Section 3.8, we address the specific problem of overlapping and touching components. Concluding remarks are given in Section 4.

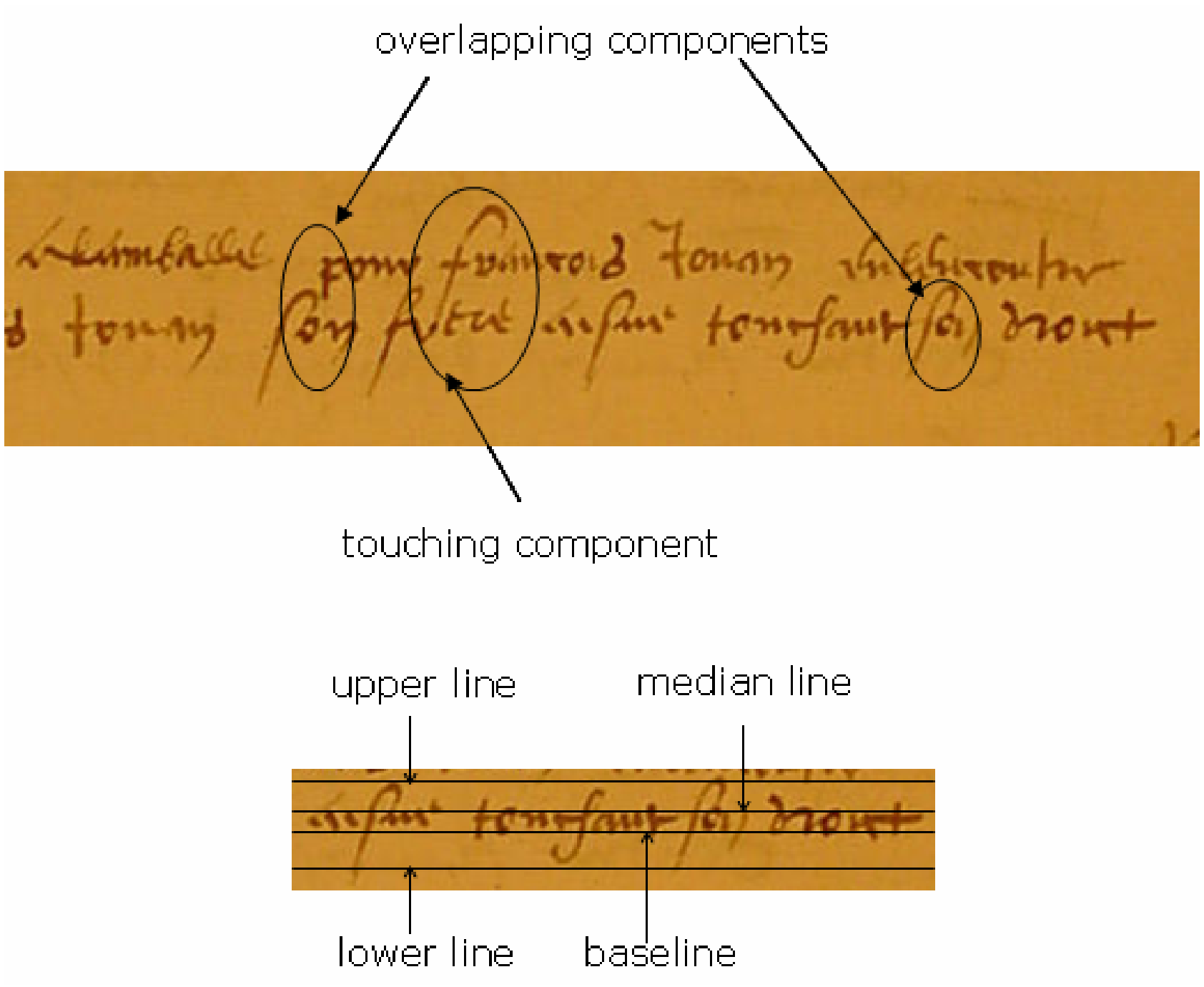

Fig. 2. Reference lines and interfering lines with overlapping and touching components.

## 2. Characteristics and representation of text lines

To have a good idea of the physical structure of a document image, one only needs to look at it from a certain distance: the lines and the blocks are immediately visible. These blocks consist of columns, annotations in margins, stanzas, etc... As blocks generally have no rectangular shape in historical documents, the text line structure becomes the dominant physical structure. We first give some definitions about text line components and text line segmentation. Then we describe the factors which make this text line segmentation hard. Finally, we describe how a text line can be represented.

### 2.1 Definitions

*baseline*: fictitious line which follows and joins the lower part of the character bodies in a text line (Fig. 2)

*median line*: fictitious line which follows and joins the upper part of the character bodies in a text line.

*upper line*: fictitious line which joins the top of ascenders.

*lower line*: fictitious line which joins the bottom of descenders.

*overlapping components:* overlapping components are descenders and ascenders located in the region of an adjacent line (Fig. 2).

*touching components*: touching components are ascenders and descenders belonging to consecutive lines which are thus connected. These components are large but hard to discriminate before text lines are known.

*text line segmentation*: text line segmentation is a labeling process which consists in assigning the same label to spatially aligned units (such as pixels, connected components or characteristic points). There are two categories of text line segmentation approaches: searching for (fictitious) separating lines or paths, or searching for aligned physical units. The choice of a segmentation technique depends on the complexity of the text line structure of the document.

### 2.2 Influence of author style

*baseline fluctuation*: the baseline may vary due to writer movement. It may be straight, straight by segments, or curved.

*line orientations*: there may be different line orientations, especially on authorial works where there are corrections and annotations.

*line spacing*: lines that are rather widely spaced lines are easy to find. The process of extracting text lines grows more difficult as interlines are narrowing; the lower baseline of the first line is becoming closer to the upper baseline of the second line; also, descenders and ascenders start to fill the blank space left for separating two adjacent text lines (Fig. 3).

*insertions*: words or short text lines may appear between the principal text lines, or in the margins.

### 2.3 Influence of poor image quality

*imperfect preprocessing*: smudges, variable background intensity and the presence of seeping ink from the other side of the document make image preprocessing particularly difficult and produce binarization errors.

*stroke fragmentation and merging:* punctuation, dots and broken strokes due to low-quality images and/or binarization may produce many connected components; conversely, words, characters and strokes may be split into several connected components. The broken components are no longer linked to the median baseline of the writing and become ambiguous and hard to segment into the correct text line (Fig. 3).

### 2.4 Text line representation

*separating paths and delimited strip:* separating lines (or paths) are continuous fictitious lines which can be uniformly straight, made of straight segments, or of curving joined strokes. The delimited strip between two consecutive separating lines receives the same text line label. So the text line can be represented by a strip with its couple of separating lines (Fig. 4).

*clusters:* clusters are a general set-based way of defining text lines. A label is associated with each cluster. Units within the same cluster belong to the same text line. They may be pixels, connected components, or blocks enclosing pieces of writing. A text line can be represented by a list of units with the same label.

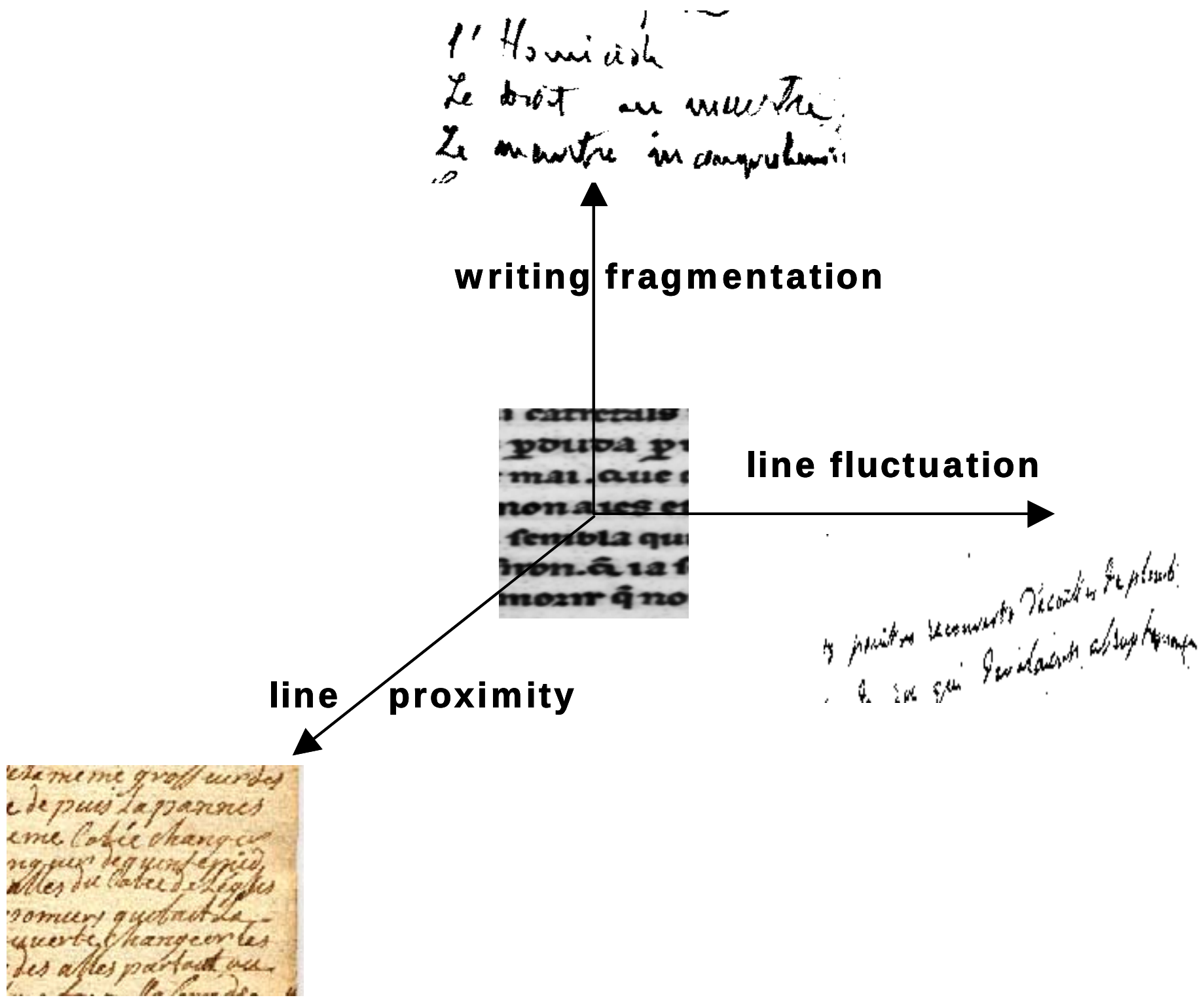

Fig. 3. The three main axes of document complexity for text line segmentation.

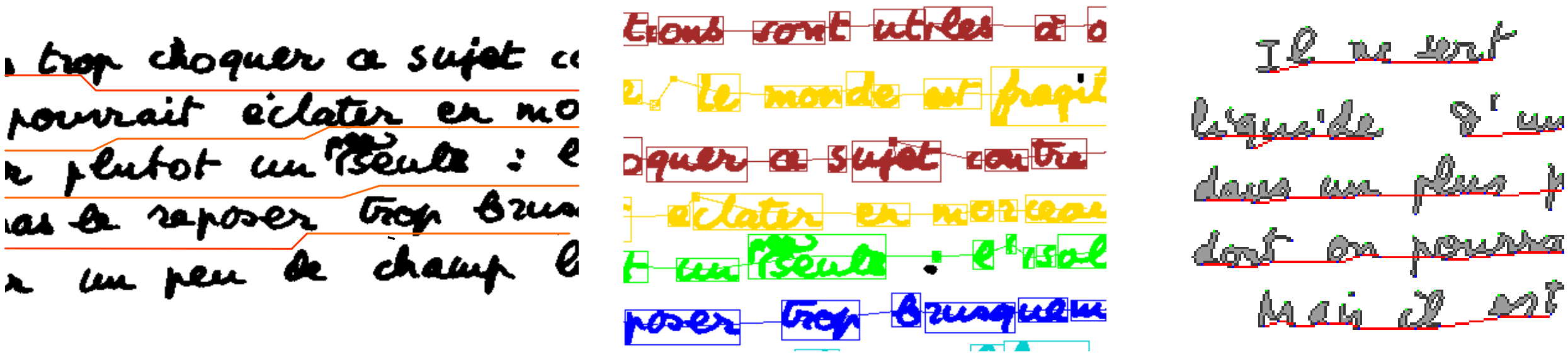

Fig. 4. Various text line representations: paths, strings and baselines.

*strings:* strings are lists of spatially aligned and ordered units. Each string represents one text line.

*baselines*: baselines follow line fluctuations but partially define a text line. Units connected to a baseline are assumed to belong to it. Complementary processing has to be done to cluster non-connected units and touching components.

# 3. Text line segmentation

Printed historical documents belong to a large period from $16^{th}$ to $20^{th}$ centuries (reports, ancient books, registers, card archives). Their printing may be faint, producing writing fragmentation artifacts. However, text lines are still enclosed in rectangular areas. After the text part has been extracted and restored, top-down and smearing techniques are generally applied for text line segmentation. A large proportion of historical documents are handwritten: scrolls, registers, private and official letters, authorial drafts. The type of writing differs considerably from one document to another. It can be calligraphed or cursive; various styles can be observed (Fig. 1). In the context of cursive handwriting, statistical information about line spacing and line orientation is hard to capture. Several techniques, which take into account handwriting and layout irregularities, as well as local and global characteristics of the text lines, have been developed

## 3.1 Preprocessing

Text line extraction would ideally process document images without background noise and without non-textual elements; the writing would be well contrasted with as little fragmentation as possible. In reality, preprocessing is often necessary. Although preprocessing has to be accurately adapted to each document and to its characteristics, we shortly describe here some preprocessing techniques that can be performed before text line extraction.

Non-textual elements around the text such as book bindings, book sides, parts of fingers (thumb marks from someone holding the book open f.i.) should be removed upon criteria such as position and intensity level. On the document itself, holes, stains, may be removed by high-pass filtering [12]. Other non-textual elements (stamps, seals) but also ornamentation, decorated initials, can be removed using knowledge about the shape, the color or the position of these elements [17]. Extracting text from figures (text segmentation) can also be performed on texture grounds [20][36] or by morphological filters [16][37]. Linear graphical elements such as big crosses (called “St Andre’s crosses”) appear in some of Flaubert’s manuscripts. Removing these elements is performed through GUI by Kalman filtering in [31].

Textual but unwanted elements such as the writing on the verso (bleed through text) may be removed by filtering and wavelet techniques [24][54][32] and by combining the verso image (the reverse side image) with the recto one (front side image).

Binarization, if necessary, can be performed by global or local thresholding. Global thresholding algorithms are not generally applicable to historical documents, due to inhomogeneous background. Thus, global thresholding results in severe deterioration in the quality of the segmented document image. Several local thresholding techniques have already been proposed to partially overcome such difficulties [21]. These local methods determine the threshold values based on the local properties of an image, e.g. pixel-by-pixel or region-by-region, and yield relatively better binarization results when compared with global thresholding methods. Writing may be faint so that over-segmentation or under-segmentation may occur. The integral ratio technique [52] is a two-stage segmentation technique adapted to this problem. Background normalization [51] can be performed before binarization in order to find a global threshold more easily.

### 3.2 Projection–based methods

Projection-profiles are commonly used for printed document segmentation. This technique can also be adapted to handwritten documents with little overlap. The vertical projection-profile is obtained by summing pixel values along the horizontal axis for each y value. From the vertical profile, the gaps between the text lines in the vertical direction can be observed (Fig. 5).

$$profile(y) = \sum_{1 \le x \le M} f(x, y)$$

The vertical profile is not sensitive to writing fragmentation. Variants for obtaining a profile curve may consist in projecting black/white transitions such as in Marti and Bunke [35] or the number of connected components, rather than pixels. The profile curve can be smoothed, e.g. by a Gaussian or median filter to eliminate local maxima [33]. The profile curve is then analysed to find its maxima and minima. There are two drawbacks: short lines will provide low peaks, and very narrow lines, as well as those including many overlapping components will not produce significant peaks. In case of skew or moderate fluctuations of the text lines, the image may be divided into vertical strips and profiles sought inside each strip (Zahour *et al.* [58]). These piecewise projections are thus a means of adapting to local fluctuations within a more global scheme.

In Shapiro *et al.*[49], the global orientation (skew angle) of a handwritten page is first searched by applying a Hough transform on the entire image. Once this skew angle is obtained, projections are achieved along this angle. The number of maxima of the profile give the number of lines. Low maxima are discarded on their value, which is compared to the highest maxima. Lines are delimited by strips, searching for the minima of projection profiles around each maxima. This technique has been tested on a set of 200 pages within a word segmentation task.

In the work of Antonacopoulos and Karatzas [3], each minimum of the profile curve is a potential segmentation point.  Potential points are then scored according to their distance to adjacent segmentation points. The reference distance is obtained from the histogram of distances between adjacent potential segmentation points. The highest scored segmentation point is used as an anchor to derive the remaining ones. The method is applied to printed records of the second World War which have regularly spaced text lines. The logical structure is used to derive the text regions where the names of interest can be found.

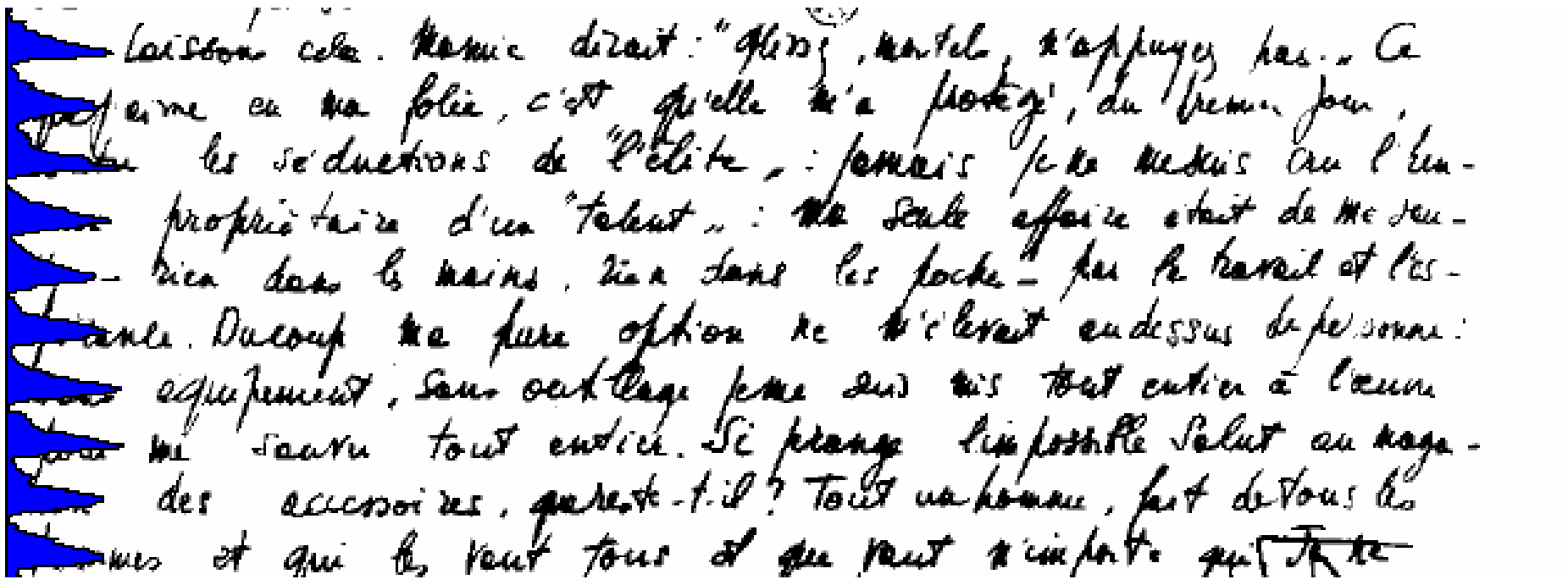

Fig. 5. Vertical projection-profile extracted on an autograph of Jean-Paul Sartre.

The RXY cuts method applied in He and Downton [18], uses alternating projections along the X and the Y axis. This results in a hierarchical tree structure. Cuts are found within white spaces. Thresholds are necessary to derive inter-line or inter-block distances. This method can be applied to printed documents (which are assumed to have these regular distances) or well separated handwritten lines.

### 3.3 Smearing methods

For printed and binarized documents, smearing methods such as the Run-Length Smoothing Algorithm (Wong *et al.* [57]) can be applied. Consecutive black pixels along the horizontal direction are smeared: i.e. the white space between them is filled with black pixels if their distance is within a predefined threshold. The bounding boxes of the connected components in the smeared image enclose text lines.

A variant of this method adapted to gray level images and applied to printed books from the sixteenth century consists in accumulating the image gradient along the horizontal direction (LeBourgeois [25]). This method has been adapted to old printed documents within the Debora project [26]. For this purpose, numerous adjustments in the method concern the tolerance for character alignment and line justification.

Text line patterns are found in the work of Shi and Govindaraju [50] by building a fuzzy run length matrix. At each pixel, the fuzzy run-length is the maximal extent of the background along the horizontal direction. Some foreground pixels may be skipped if their number does not exceed a predefined value. This matrix is threshold to make pieces of text lines appear without ascenders and descenders (Fig. 6). Parameters have to be accurately and dynamically tuned.

### 3.4 Grouping methods

These methods consist in building alignments by aggregating units in a bottom-up strategy. The units may be pixels or of higher level, such as connected components, blocks or other features such as salient points. Units are then joined together to form alignments. The joining scheme relies on both local and global criteria, which are used for checking local and global consistency respectively.

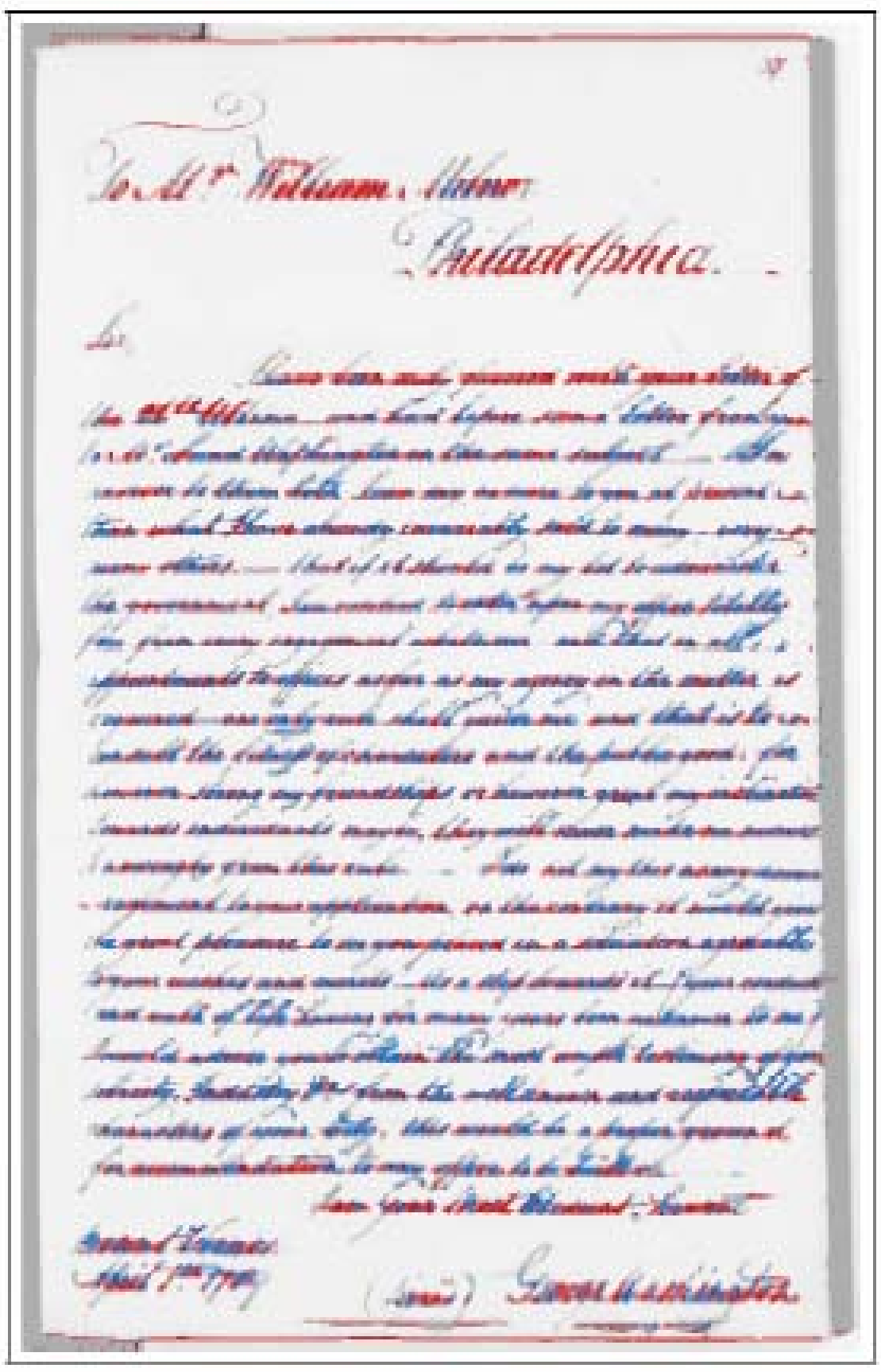

Fig. 6 Text line patterns extracted from a letter of Georges Washington (reprinted from Shi and Govindaraju [50], © [2004] IEEE). Foreground pixels have been smeared along the horizontal direction.

Contrary to printed documents, a simple nearest-neighbor joining scheme would often fail to group complex handwritten units, as the nearest neighbor often belongs to another line. The joining criteria used in the methods described below are adapted to the type of the units and the characteristics of the documents under study. But every method has to face the following:

- initiating alignments: one or several seeds for each alignment.
- defining a unit's neighborhood for reaching the next unit. It is generally a rectangular or angular area (Fig. 7).
- solving conflicts. As one unit may belong to several alignments under construction, a choice has to be made: discard one alignment or keep both of them, cutting the unit into several parts.

Hence, these methods include one or several quality measures which ensure that the text line under construction is of good quality. When comparing the quality measures of two alignments in conflict, the alignment of lower quality can be discarded (Fig. 7). Also, during the grouping process, it is possible to choose between the different units that can be aggregated within the same neighborhood by evaluating the quality of each of the so-formed alignments.

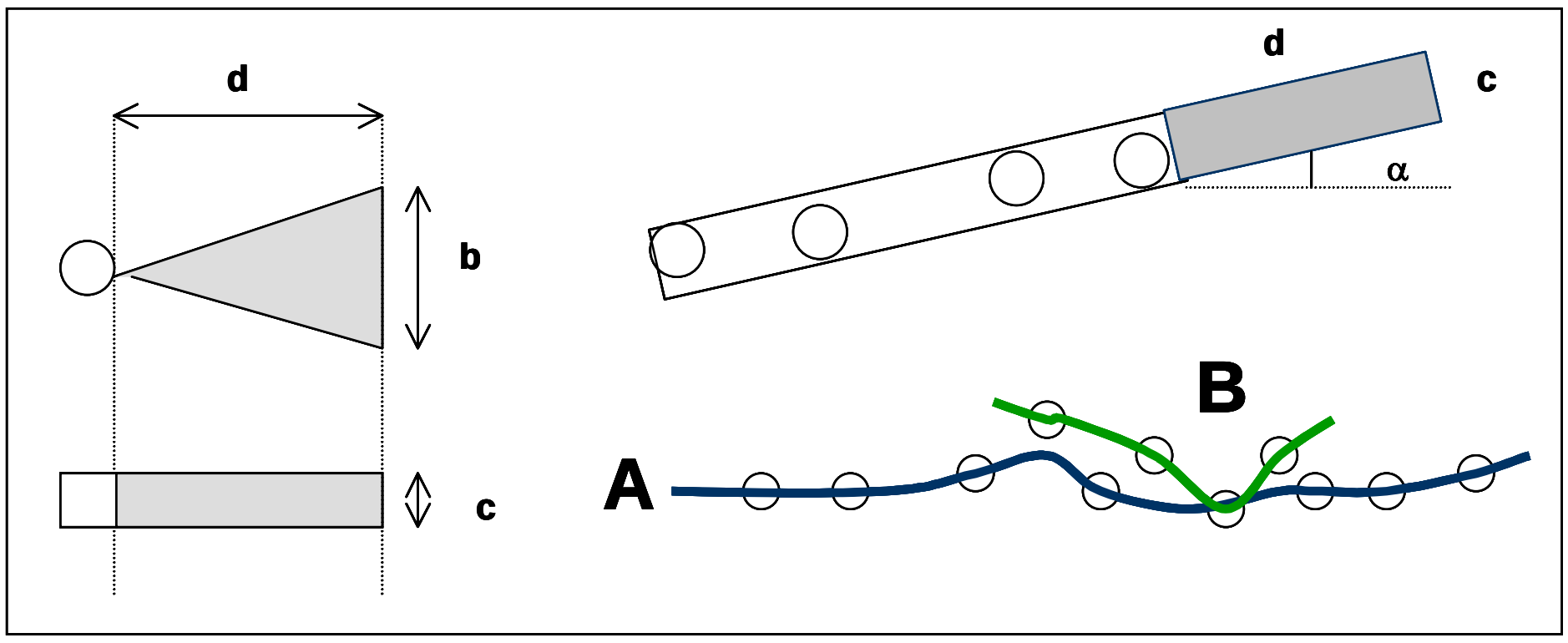

Fig. 7. Angular and rectangular neighborhoods from point and rectangular units (left). Neighborhood defined by a cluster of units (upright). Two alignments A and B in conflict: a quality measure will choose A and discard B (down right).

Quality measures generally include the *strength* of the alignment, i.e. the number of units included. Other quality elements may concern component size, component spacing, or a measure of the alignment's straightness.

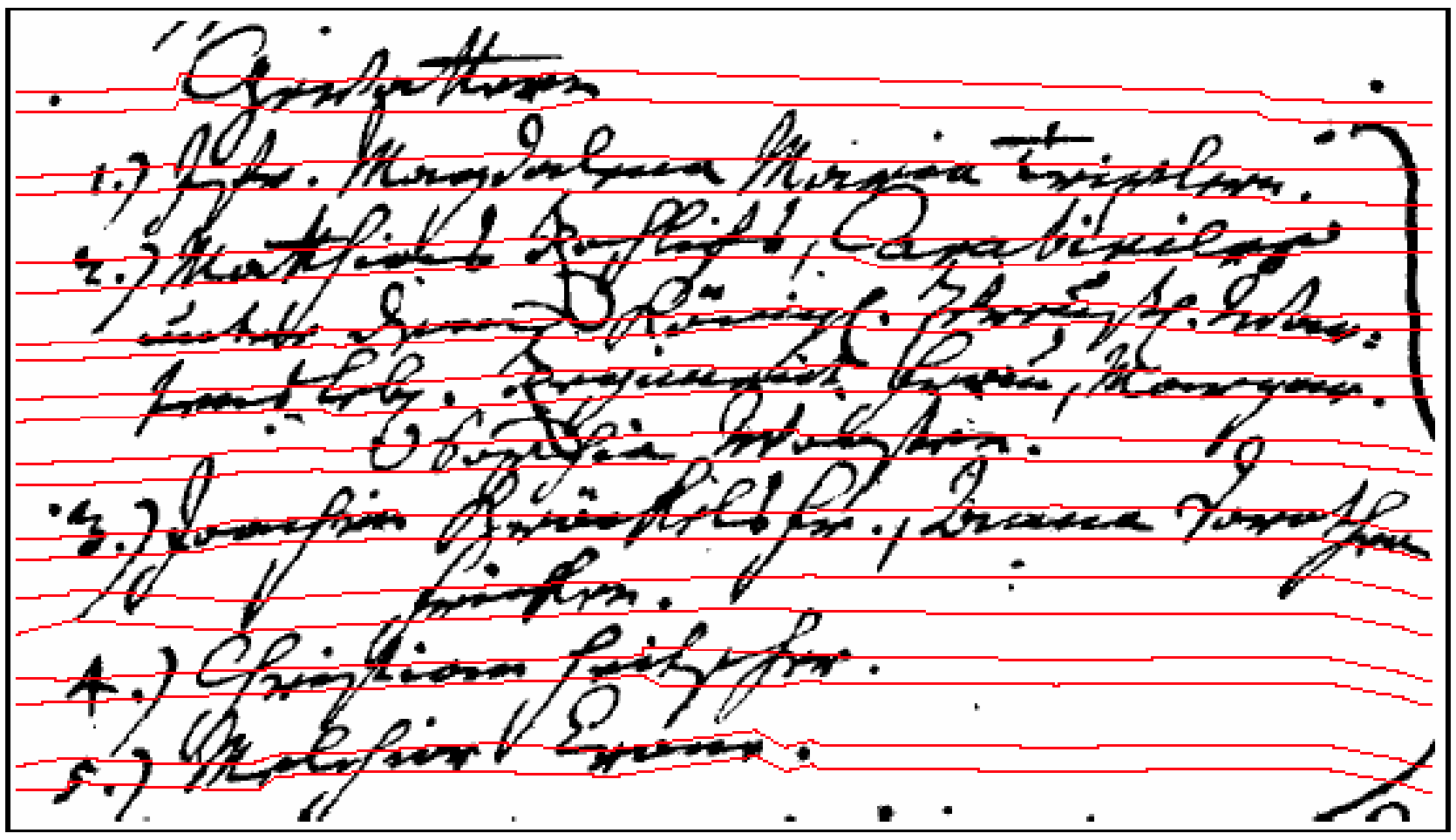

Fig. 8. Text lines extracted on Church Registers (reprinted from Feldbach [12] with permission from the author).

Likforman-Sulem and Faure have developed in [28] an iterative method based on perceptual grouping for forming alignments, which has been applied to handwritten pages, author drafts and historical documents [29][47]. Anchors are detected by selecting connected components elongated in specific directions (0°, 45°, 90°, 125°). Each of these anchors becomes the seed of an alignment. First, each anchor, then each alignment, is extended to the left and to the right. This extension uses three Gestalt criteria for grouping components: proximity, similarity and direction continuity. The threshold is iteratively incremented in order to group components within a broader neighborhood until no change occurs. Between each iteration, alignment quality is checked by a quality measure which gives higher rates to long alignments including anchors of the same direction. A penalty is given when the alignment includes anchors of different directions. Two alignments may cross each other, or overlap. A set of rules is applied to solve these conflicts taking into account the quality of each alignment and neighboring components of higher order (Fig. 14).

In the work of Feldbach and Tönnies [12][13], body baselines are searched in Church Registers images. These documents include lots of fluctuating and overlapping lines. Baselines units are the minima points of the writing (obtained here from the skeleton). First basic line segments (BLS) are constructed, joining each minima point to its neighbors. This neighborhood is defined by an angular region (+-20°) for the first unit grouped, then by a rectangular region enclosing the points already joined for the remaining ones. Unwanted basic segments are found from minima points detected in descenders and ascenders. These segments may be isolated or in conflict with others. Various heuristics are defined to eliminate alignments on their size, or the local inter-line distance and on a quality measure which favours alignments whose units are in the same direction rather than nearer units but positioned lower or higher than the current direction. Conflicting alignments can be reconstructed depending on the topology of the conflicting alignments. The median line is searched from the baseline and from maxima points (Fig. 8). Pixels lying within a given baseline and median line are clustered in the corresponding text line, while ascenders and descenders are not segmented. Correct segmentation rates are reported between 90% and 97 % with adequate parameter adjustment. The seven documents tested range from the $17^{th}$ to the $19^{th}$ century.

### 3.5 Methods based on the Hough transform

The Hough transform is a very popular technique [19] for finding straight lines in images. In Likforman-Sulem *et al.* [30], a method has been developed on a hypothesis-validation scheme. Potential alignments are hypothesized in the Hough domain and validated in the Image domain. Thus, no assumption is made about text line directions (several may exist within the same page). The centroids of the connected components are the units for the Hough transform. A set of aligned units in the image along a line with parameters ($\rho$, $\theta$) is included in the corresponding cell ($\rho$, $\theta$) of the Hough domain. Alignments including a lot of units correspond to high peaked cells of the Hough domain. To take into account fluctuations of handwritten text lines, i.e. the fact that units within a text line are not perfectly aligned, two hypotheses are considered for each alignment and an alignment is formed from units of the *cell structure* of a *primary cell.*

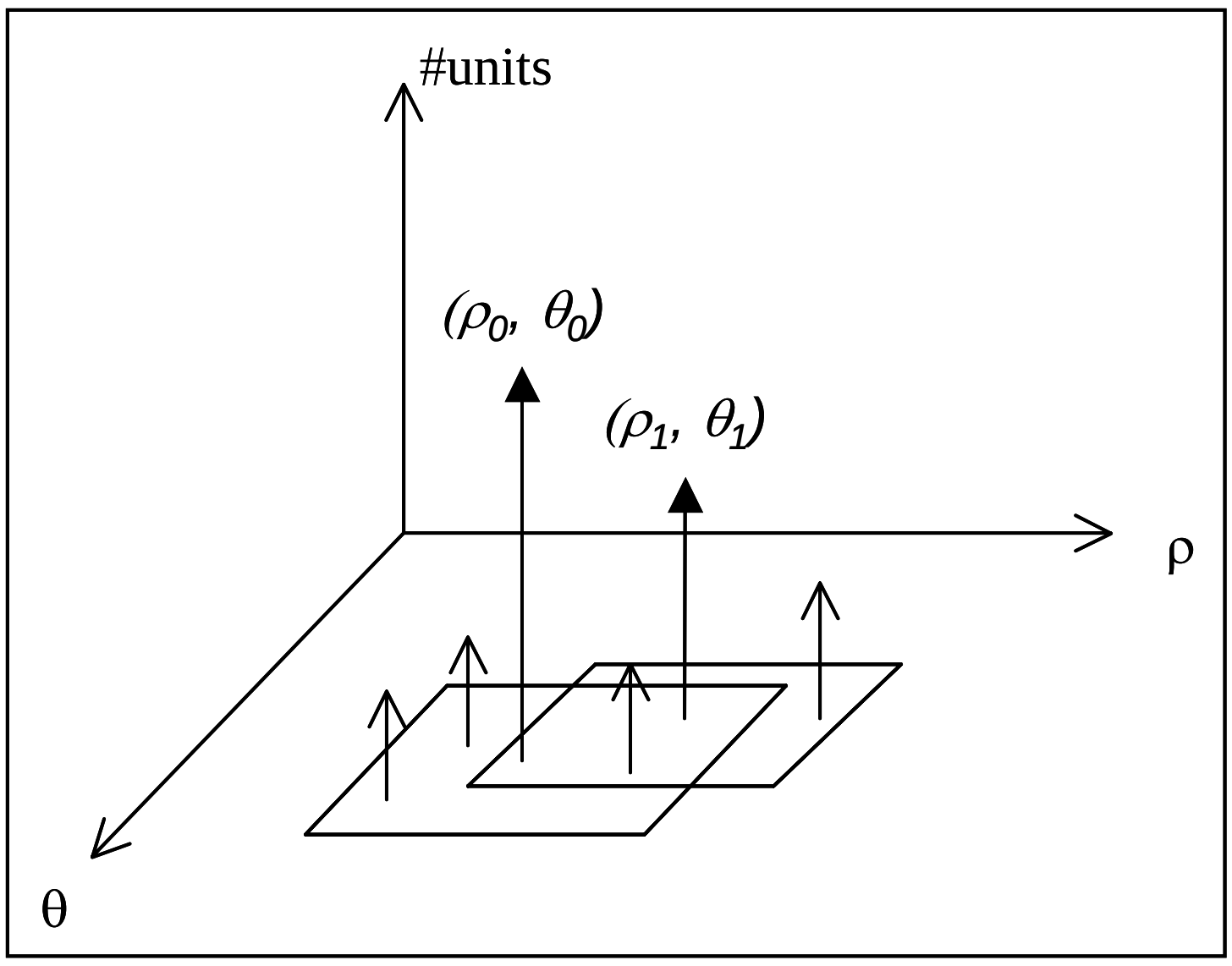

Fig. 9. Hypothesized cells ($\rho_0$, $\theta_0$) and ($\rho_1$, $\theta_1$) in Hough space. Each peak corresponds to perfectly aligned units. An alignment is composed of units belonging to a cluster of cells (the *cell structure*) around a primary cell.

A cell structure of a cell ($\rho$, $\theta$) includes all the cells lying in a cluster centered around ($\rho$, $\theta$). Consider the cell ($\rho_0$, $\theta_0$) having the greatest count of units. A second hypothesis ($\rho_1$, $\theta_1$) is searched in the cell structure of ($\rho_0$, $\theta_0$). The alignment chosen between these two hypotheses is the strongest one, i.e. the one which includes the highest number of units in its cell structure. And the corresponding cell ($\rho_0$, $\theta_0$) or ($\rho_1$, $\theta_1$) is the primary cell (Fig. 9).

However, actual text lines rarely correspond to alignments with the highest number of units as crossing alignments (from top to bottom for writing in horizontal direction) must contain more units than actual text lines. A potential alignment is validated (or invalidated) using contextual information, i.e. considering its *internal* and *external* neighbors. An internal neighbor of a unit j is a within-Hough alignment neighbor. An external neighbor is a out of Hough alignment neighbor which lies within a circle of radius $\delta_j$ from unit j. Distance $\delta_j$ is the average distance of the internal neighbor distances from unit j. To be validated, a potential alignment may contain fewer external units than internal ones. This enables the rejection of alignments which have no perceptual relevance. This method can extract oriented text lines and sloped annotations under the assumption that such lines are almost straight (Fig. 10).

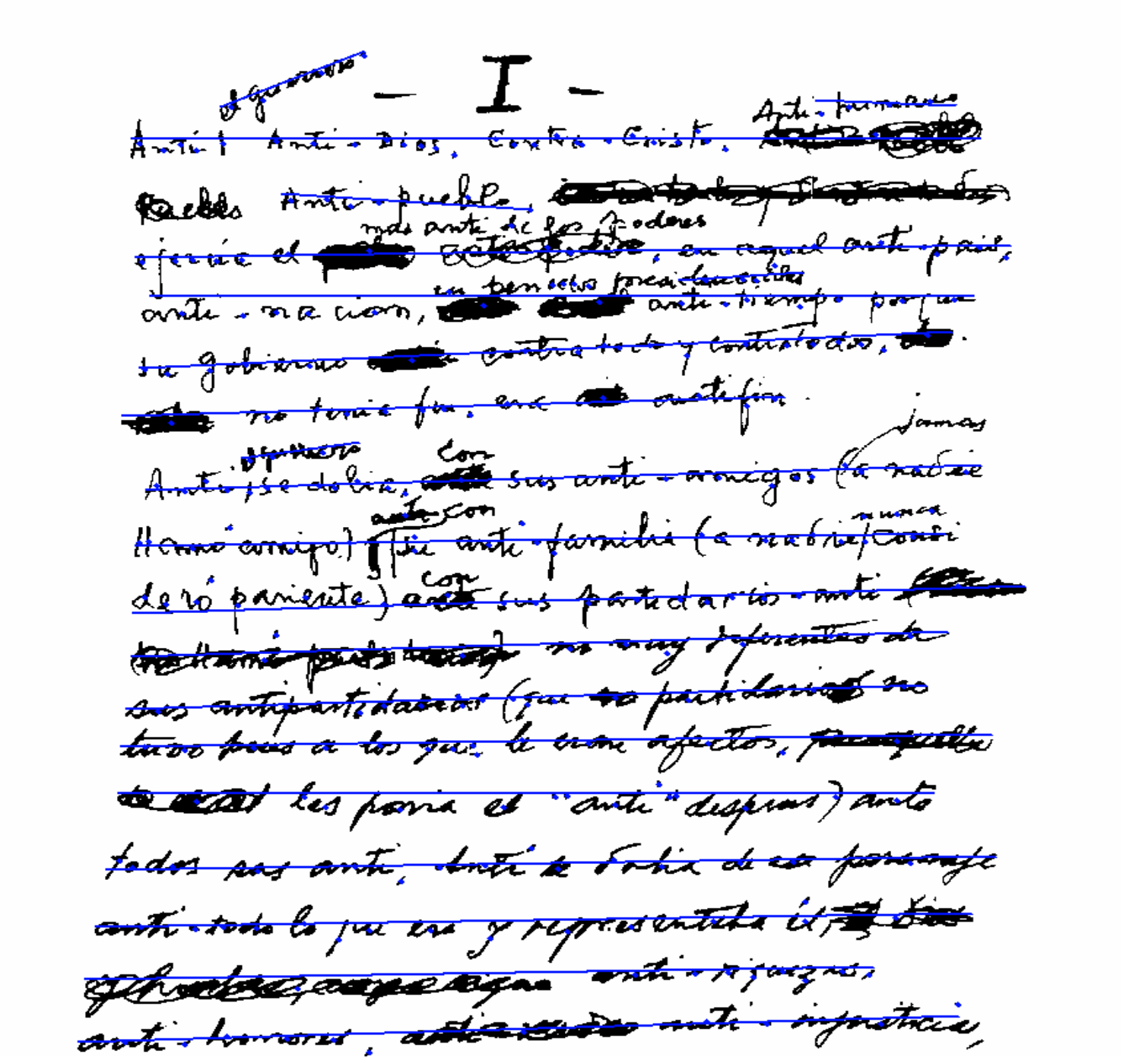

Fig. 10. Text lines extracted on an autograph of Miguel Angel Asturias. The orientations of traced lines correspond to those of the primary cells found in Hough space.

The Hough transform can also be applied to fluctuating lines of handwritten drafts such as in Pu and Shi [45]. The Hough transform is first applied to minima points (units) in a vertical strip on the left of the image. The alignments in the Hough domain are searched starting from a main direction, by grouping cells in an exhaustive search in 6 directions. Then a moving window, associated with a clustering scheme in the image domain, assigns the remaining units to alignments. The clustering scheme (*Natural Learning Algorithm*) allows the creation of new lines starting in the middle of the page.

### 3.6 Repulsive-Attractive network method

An approach based on *attractive-repulsive forces* is presented in Oztop *et al.* [40]. It works directly on grey-level images and consists in iteratively adapting the y-position of a predefined number of baseline units. Baselines are constructed one by one from the top of the image to bottom. Pixels of the image act as attractive forces for baselines and already extracted baselines act as repulsive forces. The baseline to extract is initialized just under the previously examined one, in order to be repelled by it and attracted by the pixels of the line below (the first one is initialized in the blank space at top of the document). The lines must have similar lengths. The result is a set of pseudo-baselines, each one passing through word bodies (Fig. 11). The method is applied to ancient Ottoman document archives and Latin texts.

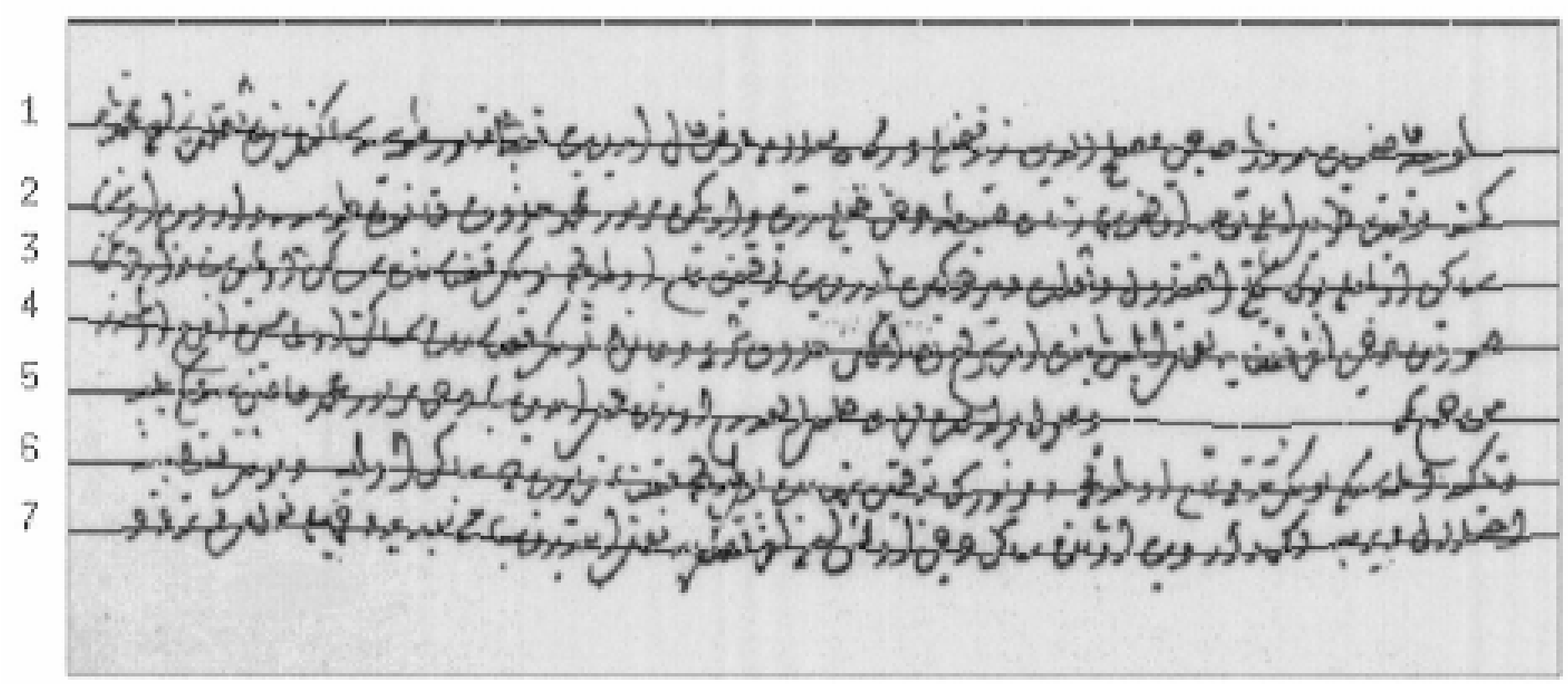

Fig. 11. Pseudo baselines extracted by a Repulsive-Attractive network on an Ancient Ottoman text (reprinted from Oztop *et al.* [40] Copyright (1999) with permission from Elsevier).

### 3.7 Stochastic method

We present here a method based on a *probabilistic Viterbi* algorithm (Tseng and Lee [56]), which derives non-linear paths between overlapping text lines. Although this method has been applied to modern Chinese handwritten documents, this principle could be enlarged to historical documents which often include overlapping lines. Lines are extracted through hidden Markov modeling. The image is first divided into little cells (depending on stroke width), each one corresponding to a state of the HMM (Hidden Markov Model). The best segmentation paths are searched from left to right; they correspond to paths which do not cross lots of black points and which are as straight as possible. However, the displacement in the graph is limited to immediately superior or inferior grids. All best paths ending at each y location of the image are considered first. Elimination of some of these paths uses a quality threshold T: a path whose probability is less than T is discarded. Shifted paths are easily eliminated (and close paths are removed on quality criteria). The method succeeds when the ground truth path between text lines is slightly changing along the y-direction (Fig. 12). In the case of touching components, the path of highest probability will cross the touching component at points with as less black pixels as possible. But the method may fail if the contact point contains a lot of black pixels.

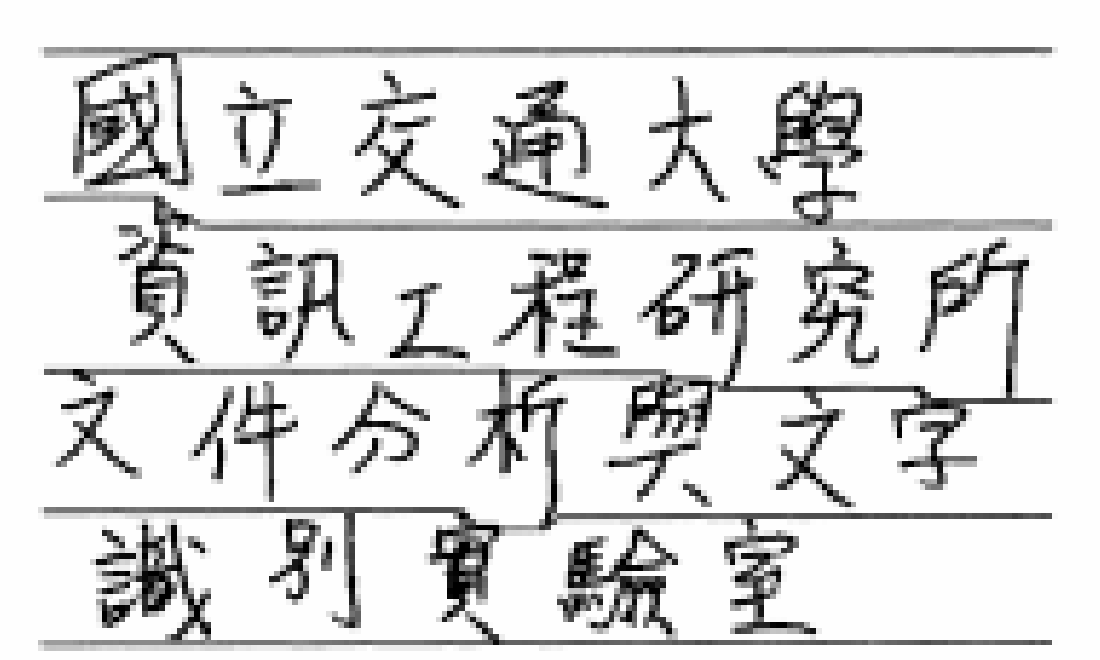

Fig. 12. Segmentation paths obtained by a stochastic method (reprinted from Tseng and Lee [56], Copyright (1999) with permission from Elsevier).

### 3.8 Processing of overlapping and touching components

Overlapping and touching components are the main challenges for text line extractions since no white space is left between lines. Some of the methods surveyed above do not need to detect such components because they extract only baselines (3.4, 3.6), or because in the method itself some criteria make paths avoid crossing black pixels (c.f. Section 3.7). This section only deals with methods where ambiguous components (overlapping or touching) are actually detected before, during or after text line segmentation

Such criteria as component size, the fact that the component belongs to several alignments, or on the contrary to no alignment, can be used for detecting ambiguous components. Once the component is detected as ambiguous, it must be classified into three categories: the component is an overlapping component which belongs to the upper (resp. lower) alignment, the component is a touching component which has to be decomposed into several parts (two or more parts, as components may belong to three or more alignments in historical documents). The separation along the vertical direction is a hard problem which can be done roughly (horizontal cut), or more accurately by analysing stroke contours and referring to typical configurations (Fig. 13).

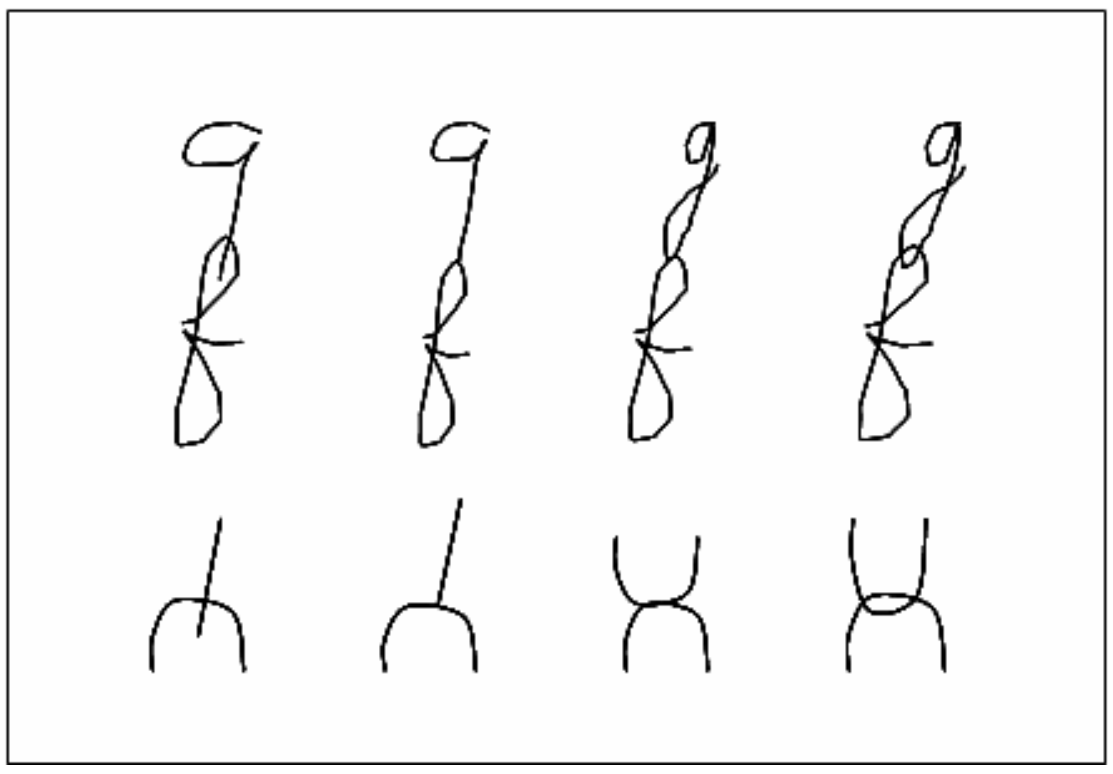

Fig. 13. Set of typical overlapping configurations (adapted from Piquin *et al.* [44]).

The grouping technique presented in Section 3.4 detects an ambiguous component during the grouping process when a conflict occurs between two alignments [28] [29]. A set of rules is applied to label the component as overlapping or touching. The ambiguous component extends in each alignment region. The rules use as features the density of black pixels of the component in each alignment region, alignment proximity and contextual information (positions of both alignments around the component). An overlapping component will be assigned to only one alignment. And the separation of a touching component is roughly performed by drawing a horizontal frontier segment. The frontier segment position is decided by analysing the vertical projection profile of the component. If the projection profile includes two peaks, the cut will be done middle way from them, as in Figure 14. Else the component will be cut into two equal parts.

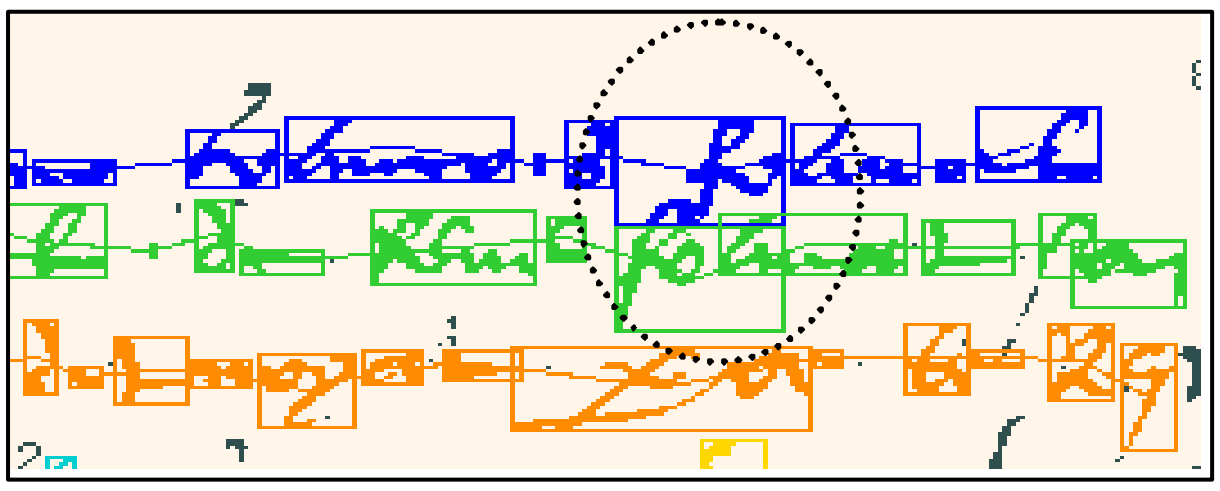

Fig. 14. Touching component separated in a ‘Lettre de Remission’.

In Likforman-Sulem *et al.* [30], touching and overlapping components are detected after the text line extraction process described in Section 3.5. These components are those which are intersected by at least two different lines (ρ,θ) corresponding to primary cells of validated alignments.

In Zahour *et al.* [58][59], the document page is first cut into eight equal columns. A projection-profile is performed on each column. In each histogram, two consecutive minima delimit a text block. In order to detect touching and overlapping components, a k-means clustering scheme is used to classify the text blocks so extracted into three classes: big, average, small. Overlapping components necessarily belong to big physical blocks. All the overlapping cases are found in the big text blocks class. All the “one line” blocks are grouped in the average block text class. A second k-means clustering scheme finds the actual inter-line blocks; put together with the “one line” block size, this determines the number of pieces a large text block must be cut into (cf. Fig. 16).

A similar method such as the one presented above is applied to Bangla handwriting Indian documents in Pal and Datta [41]. The document is divided into vertical strips. Profile cuts within each strip are computed to obtain anchor points of segmentation (PSLs) which do not cross any black pixels. These points are grouped through strips by neighboring criteria. If no segmentation point is present in the adjacent strip, the baseline is extended near the first black pixel encountered which belongs to an overlapping or touching component. This component is classified as overlapping or touching by analysing its vertical extension (upper, lower) from each side of the intersection point. An empirical rule classifies the component. In the touching case, the component is horizontally cut at the intersection point (Fig. 15).

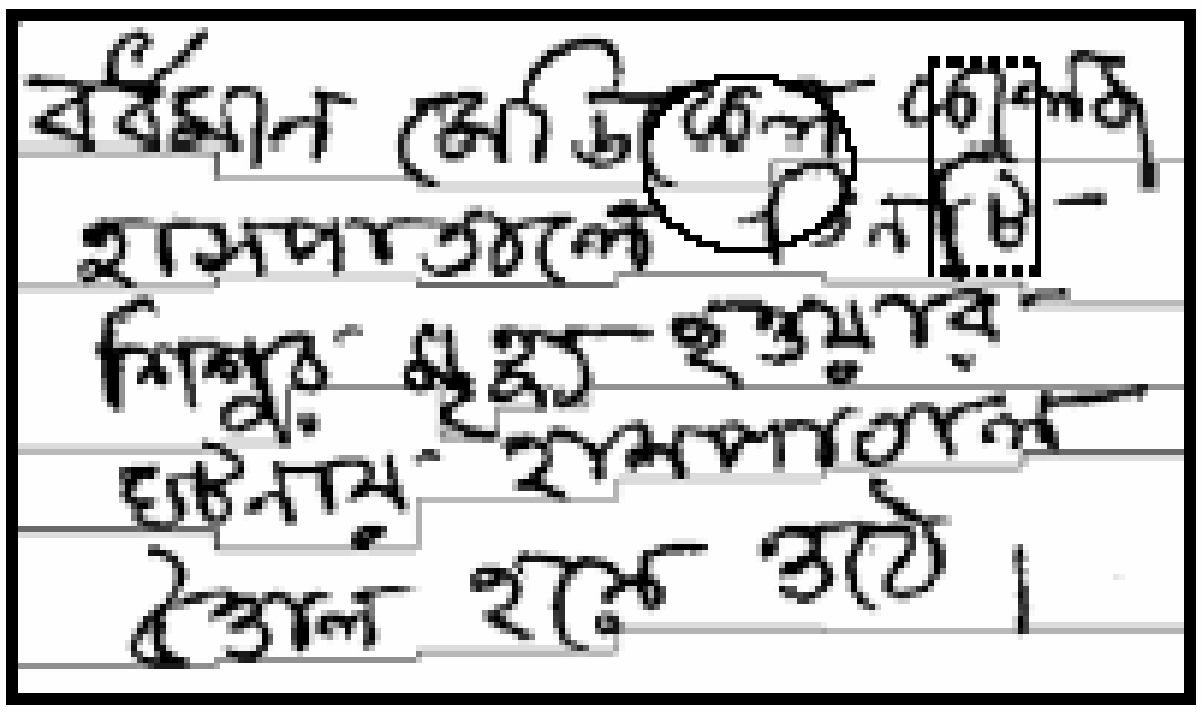

Fig.15. Overlapping components separated (circle) and touching component separated into two parts (rectangle) in Bangla writing (from Pal and Datta [41], © [2003] IEEE).

Some solutions for separation of units belonging to several text lines can be found also in the case of mail pieces and handwritten databases where efforts have been made for recognition purposes [44] [7]. In the work of Piquin *et al.* [44], separation is made from the skeleton of touching characters and the use of a dictionary of possible touching configurations (Fig. 13). In Bruzzone and Coffetti [7], the contact point between ambiguous strokes is detected and processed from their external border. An accurate analysis of the contour near the contact point is performed in order to separate the strokes according to two registered configurations: a loop in contact with a stroke, or two loops in contact. In simple cases of handwritten pages (Marti and Bunke [35]), the center of gravity of the connected component is used either to associate the component to the current line or to the following line, or to cut the component into two parts. This works well if the component is a single character. It may fail if the component is a word, or part of a word, or even several words.

## 3.9 Non Latin documents

The inter-line space in Latin documents is filled with single dots, ascenders and descenders. The Arabic script is connected and cursive. Large loops are present in the inter-line space and ancient Arabic documents include diacritical points [1]. In the Hebrew squared writing, the baseline is situated on top of characters. Documents such as decorated Bibles, prayer books and scientific treatises include diacritical points which represent vowels. Ancient Hebrew documents may include decorated words but no decorated initials as there is no upper/lower case character concept in this script. In the alphabets of some Indian scripts (like Devnagari, Bangla and Gurumukhi), many basic characters have an horizontal line (the head line) in the upper part [42]. In Bangla and Telugu text, touching and overlapping occur frequently [23].To date, the published studies on historical documents concern Arabic and Hebrew. Work about Chinese and Bangla Indian writings on good quality documents have been already mentioned in Sections 3.7 and 3.8: they should be also suitable to ancient documents as they include local processing.

### *3.9.1 Ancient Arabic documents*

Figure 1 is a handwritten page extracted from a book of the Tunisian National Library. The writing is dense and inter-line space is faint. Several consecutive lines are often connected by one character at least, and the overlapping situations are obvious. Baseline waving produces various text orientations.

The method developed in Zahour *et al.* [59] begins with the detection of overlapping and touching components presented in §3.8, using a two-stage clustering process which separates big blocks including several lines into several parts. Blocks are then linked by neighborhood using the y coordinates. Figure 16 shows line separators using the clustering technique recursively, as described in Section 3.8.

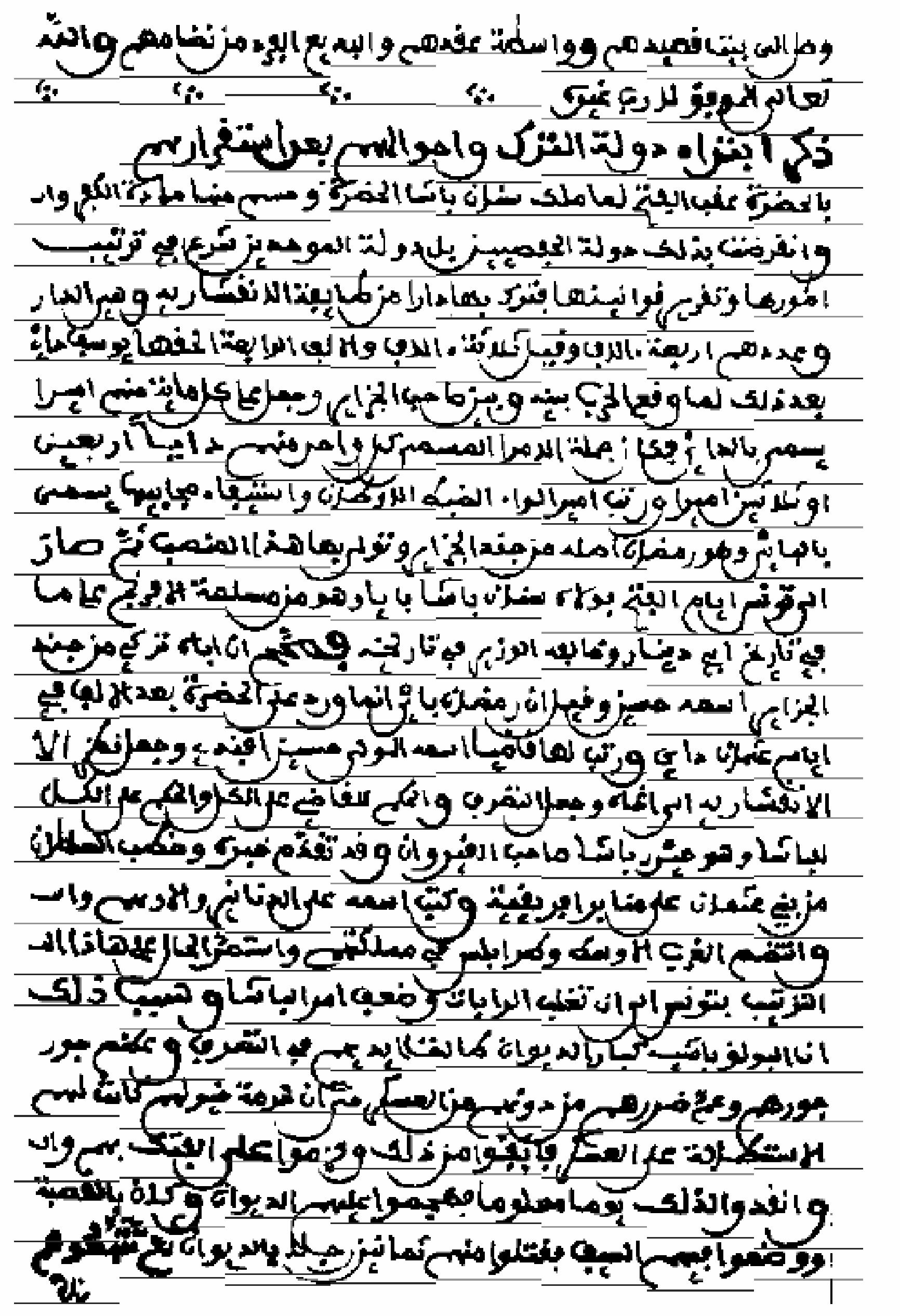

Fig. 16. Text line segmentation of the ancient Arabic handwritten document in Fig. 1.

### 3.9.1 Ancient Hebrew documents

The manuscripts studied in Likforman-Sulem *et al.* [27], are written in Hebrew, in a so-called squared writing as most characters are made of horizontal and vertical strokes. They are reproducing the biblical text of the Pentateuch. Characters are calligraphed by skilled scribes with a quill or a calamus. The Scrolls, intended to be used in the synagogue, do not include diacritics. Characters and words are written properly separated but digitization make some characters touch. Cases of overlapping components occur as characters such as Lamed, Kaf, and final letters include ascenders and descenders. Since the majority of characters are composed of one connected component, it is more convenient to perform text line segmentation from connected components units. Fig. 17 shows the resulting segmentation with the Hough-based method presented in Section 3.5.

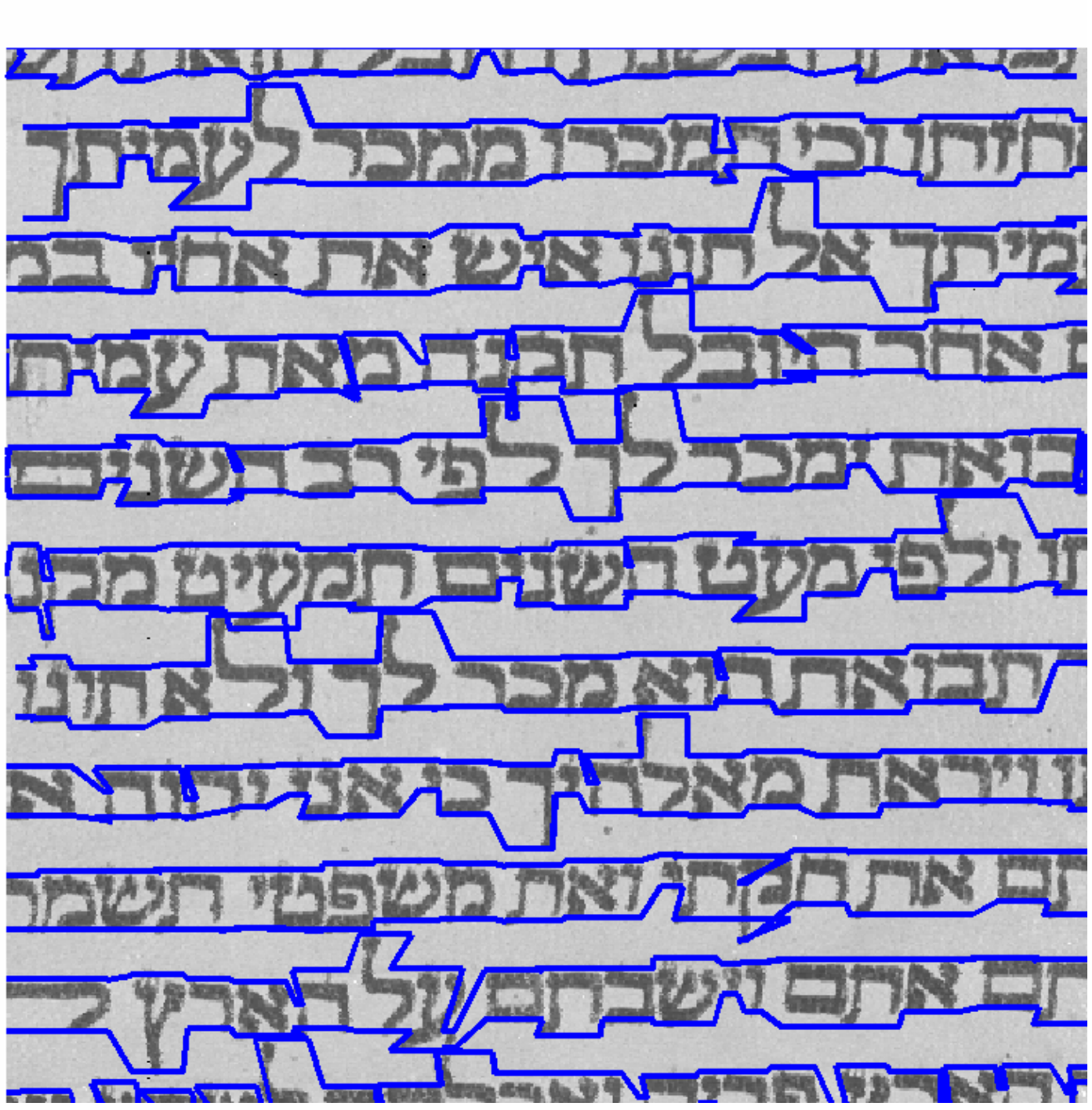

Fig. 17. Text line segmentation of a Hebrew document (Scroll).

Table 1. Text line segmentation methods suitable for historical documents

| Authors | Description | Line Description | Writing Type | Units | Suitable for | Project/ Documents |
|---|---|---|---|---|---|---|
| [Antonacopoulos and Karatzas, 2004] | projection profiles | linear paths | Latin printed | pixels | separated lines | Memorial/personal records(World War II) |
| [Calabretto and Bozzi, 1998] | projection profiles (gray level image) | linear paths | cursive handwriting | pixels | separated lines | Bambi/italian manuscripts ($16^{th}$ century) |
| [Feldbach and Tönnies, 2001] | grouping method | baselines | cursive handwriting | minima points | fluctuating lines | Church registers ($18^{th}$, $19^{th}$ century) |
| [He and Downton, 2003] | projections (RXY cuts) | linear paths | Latin printed and handwriting | pixels | separated lines | Viadocs/ Natural History Cards |
| [Lebourgeois *et al.*, 2001] | smearing (accumulated gradients) | clusters | Latin printed | pixels | separated lines | Debora/books ($16^{th}$ century) |
| [Likforman-Sulem and Faure, 1994] | grouping | strings | Latin handwriting | connected components | fluctuating lines | Philectre/ authorial manuscripts |
| [Likforman-Sulem *et al.*, 1995] | Hough transform, (hypothesis-validation scheme) | strings | Latin handwriting | connected components | different straight line directions | Philectre/ authorial manuscripts, manuscripts of the $16^{th}$ century |
| [Oztop *et al.*, 1997] | repulsive - attractive network | baselines | Arabic and Latin handwriting | pixels (gray levels) | fluctuating lines (same size) | ancient Ottoman documents |
| [Pal and Datta, 2003] | piecewise projections | piecewise linear paths | Bangla handwriting | segmentation points | overlapping/ touching lines | Indian handwritten documents |
| [Pu and Shi, 1998] | Hough transform (moving window) | clusters | Latin handwriting | minima points | fluctuating lines | handwritten documents |
| [Shapiro *et al.*, 1993] | projection profiles | linear paths | Latin handwriting | pixels | skewed separated lines | handwritten documents |
| [Shi and Govindaraju, 2004] | smearing (fuzzy run length) | cluster | Latin handwriting | pixels | straight touching lines | Newton, Galileo manuscripts |
| [Tseng and Lee, 1999 ] | stochastic (probabilistic Viterbi algorithm) | non linear paths | Chinese handwriting | pixels | overlapping lines | handwritten documents |
| [Zahour *et al.*, 2004] | piecewise projection and k-means clustering | piecewise linear paths | Arabic handwriting | text blocks | overlapping/ touching lines. | ancient Arabic documents |

# 4. Discussion and concluding remarks

An overview of text line segmentation methods developed within different projects is presented in Table 1. The achieved taxonomy consists in six major categories. They are listed as: projection-based, smearing, grouping, Hough-based, repulsive-attractive network and stochastic methods. Most of these methods are able to face some image degradations and writing irregularities specific to historical documents, as shown in the last column of Table 1.

Projection, smearing and Hough-based methods, classically adapted to straight lines and easier to implement, had to be completed and enriched by local considerations (piecewise projections, clustering in Hough space, use of a moving window, ascender and descender skipping), so as to solve some problems including: line proximity, overlapping or even touching strokes, fluctuating close lines, shape fragmentation occurrences. The stochastic method (achieved by the Viterbi decision algorithm) is conceptually more robust, but its implementation requires great care, particularly the initialization phase. As a matter of fact, text-line images are initially divided into mxn grids (each cell being a node), where the values of the critical parameters m and n are to be determined according to the estimated average stroke width in the images. Representing a text line by one or more baselines (RA method, minima point grouping) must be completed by labeling those pixels not connected to, or between the extracted baselines. The recurrent nature of the repulsive-attractive method may induce cascading detecting errors following a unique false or bad line extraction.

Projection and Hough-based methods are suitable for clearly separated lines. Projection-based methods can cope with few overlapping or touching components, as long text lines smooth both noise and overlapping effects. Even in more critical cases, classifying the set of blocks into “one line width” blocks and “several lines width” blocks allows the segmentation process to get statistical measures so as to segment more surely the “several lines width” blocks. As a result, the linear separator path may cross overlapping components. However, more accurate segmentation of the overlapping components can be performed after getting the global or piecewise straight separator, by looking closely at the so crossed strokes. The stochastic method naturally avoids crossing overlapping components (if they are not too close): the resulting non linear paths turn around obstacles. When lines are very close, grouping methods encounter a lot of conflicting configurations. A wrong decision in an early stage of the grouping results in errors or incomplete alignments. In case of touching components, making an accurate segmentation requires additional knowledge (compiled in a dictionary of possible configurations or represented by logical or fuzzy rules).

Concerning text line fluctuations, baseline-based representations seem to fit naturally. Methods using straight line-based representations must be modified as previously to give non linear results (by piecewise projections or neighboring considerations in Hough space). The more fluctuating the text line, the more refined local criteria must be. Accurate locally oriented processing and careful grouping rules make smearing and grouping methods convenient. The stochastic methods also seem suited, for they can generate non linear segmentation paths to separate overlapping characters, and even more to derive non linear cutting paths from touching characters by identifying the shortest paths.

Pixel based methods are naturally robust at dealing with writing fragmentation. But, as a consequence of writing fragmentation, when units become fragmented, sub-units may be located far from the baseline. Spurious characteristic points are then generated, disturbing alignment and implying a loss of accuracy, or more, a wrong final representation.

Quantitative assessment of performance is not generally yielded by the authors of the methods; when it is given, this is on a reduced set of documents. As for all segmentation methods, ground truth data are harder to obtain than for classification methods. For instance the ground truth for the real baseline may be hard to assess. Text line segmentation is often a step in the recognition algorithm and the segmentation task is not evaluated in isolation. To date, no general study has been carried out to compare the different methods. Text line representations differ and methods are generally tuned to a class of documents.

Analysis of historical document images is a relatively new domain. Text line segmentation methods have been developed within several projects which perform transcript mapping, authentication, word mapping or word recognition. As the need for recognition and mapping of handwritten material increases, text line segmentation will be used more and more. Contrary to printed modern documents, a historical document has unique characteristics due to style, artistic effect and writer skills. There is no universal segmentation method which can fit all these documents. The techniques presented here have been proposed to segment particular sets of documents. They can however be generalized to other documents with similar characteristics, with parameter tuning that depends on script size, stroke width and average spacing.

The major difficulty consists in obtaining a precise text line, with all descenders and ascenders segmented for accessing isolated words. As segmentation and recognition are dependent tasks, the exact segmentation of touching pieces of writing may need some recognition, or knowledge about the possible touching configurations. Text line segmentation algorithms will benefit from automatic extraction of document characteristics leading to an easier adaptation to the document under study.